\documentclass[runningheads]{llncs}
\usepackage{hyperref}
\usepackage{authblk}

\date{} 
\usepackage[T1]{fontenc}
\usepackage{graphicx}
\usepackage{booktabs}
\usepackage{adjustbox} 
\usepackage{subcaption} 
\usepackage{microtype} 
\usepackage{graphicx,verbatim,xcolor}
\begin{document}
\title{ReXInTheWild: A Unified Benchmark for Medical Photograph Understanding}
\titlerunning{ReXInTheWild: Medical Photograph Benchmark}
%

\author{
Oishi Banerjee, M.S.\inst{1}\thanks{Email: oishi\_banerjee@g.harvard.edu} \and
Sung Eun Kim, M.D.\inst{1,2} \and
Alexandra N. Willauer, M.D. \inst{1,3}
Julius M. Kernbach, M.D.\inst{4} \and
Abeer Rihan Alomaish, M.D.\inst{5} \and
Reema Abdulwahab S.\ Alghamdi, M.D.\inst{5} \and
Hassan Rayhan Alomaish, M.D.\inst{5} \and
Mohammed Baharoon, B.S.\inst{1} \and
Xiaoman Zhang, Ph.D.\inst{1} \and
Julian Nicolas Acosta, M.D.\inst{1} \and
Christine Zhou, M.D. \inst{1,6}
Pranav Rajpurkar, Ph.D.\inst{1}
}
\authorrunning{Banerjee, Kim, Willauer, et al.}
\institute{
Department of Biomedical Informatics, Harvard Medical School, Boston, MA \and
National Strategic Technology Research Institute, Seoul National University Hospital, Seoul, Republic of Korea \and
Department of Medicine, Division of Gastroenterology, Massachusetts General Hospital, Boston, MA \and
Department of Neuroradiology, Heidelberg University Hospital, Heidelberg, Germany \and
King Abdulaziz Medical City, National Guard Health Affairs, Riyadh, Saudi Arabia \and
Division of Pulmonary, Critical Care, and Sleep Medicine, University of Cincinnati, Cincinnati, OH
}

\maketitle              
\begin{abstract}
Everyday photographs taken with ordinary cameras are already widely used in telemedicine and other online health conversations, yet no comprehensive benchmark evaluates whether vision–language models can interpret their medical content. Analyzing these images requires both fine-grained natural image understanding and domain-specific medical reasoning, a combination that challenges both general-purpose and specialized models. We introduce ReXInTheWild, a benchmark of 955 clinician-verified multiple-choice questions spanning seven clinical topics across 484 photographs sourced from the biomedical literature. When evaluated on ReXInTheWild, leading multimodal large language models show substantial performance variation: Gemini-3 achieves 78\% accuracy, followed by Claude Opus 4.5 (72\%) and GPT-5 (68\%), while the medical specialist model MedGemma reaches only 37\%. A systematic error analysis also reveals four categories of common errors, ranging from low-level geometric errors to high-level reasoning failures and requiring different mitigation strategies. ReXInTheWild provides a challenging, clinically grounded benchmark at the intersection of natural image understanding and medical reasoning. The dataset is available at \href{https://huggingface.co/datasets/rajpurkarlab/ReXInTheWild}{https://huggingface.co/datasets/rajpurkarlab/ReXInTheWild}.

\keywords{Patient-Generated Health Data, Domain Shift, Diagnostic Reasoning, Multimodal Large Language Models, Vision-Language Models}

\end{abstract}

\section{Introduction}

Photographs taken with ordinary cameras already play a growing role in healthcare: patients share images with telemedicine applications and online health forums \cite{choe2018harnessing,burns2015medicalselfie,shen2024optimizing}, clinicians use point-of-care photography for wound monitoring and post-operative follow-up \cite{zhang2021woundimagequality,totty2018photographtelemedicine}, and consumer devices increasingly support dermatological and ophthalmologic screening \cite{shajirat2024teledermatology,chen2023visualimpairment}. Despite this widespread applicability, benchmarking for medical photographs remains fragmented. Existing benchmarks overwhelmingly address specialized imaging modalities such as X-rays and histopathology slide images \cite{lau2018radiologyvqa,he2020pathvqa,liu2021slake}, while photographic datasets have focused on dermatology \cite{Yim_DermaVQA_MICCAI2024,zeng2025mmskinenhancingdermatologyvisionlanguage} or on narrow tasks such as gait analysis \cite{zhou2024gaitpatternsbiomarkersvideobased,zafra2025healthgait,bandini2021facialmotion}. To our knowledge, no unified evaluation exists to test whether models can reason about the broad range of medical content that appears in everyday photographs, posing a unique dual challenge for current multimodal large language models (MLLM). Specialized medical models face a domain shift from controlled clinical imaging to unstandardized natural scenes, while general-purpose vision–language models may interpret images more accurately but lack the clinical knowledge to draw correct medical conclusions. Compounding both challenges, photographs taken ``in the wild" by non-clinicians vary widely in lighting, angle, and composition, making it harder to detect the fine-grained visual details that medical reasoning demands.

We introduce ReXInTheWild, a benchmark of 955 clinician-verified multiple-choice questions derived from 484 medical photographs spanning seven clinical topics — from musculoskeletal assessment to ophthalmologic and surgical evaluation. Questions were generated through a multi-stage pipeline combining automated candidate generation, adversarial distractor refinement, and two rounds of expert clinical review. We evaluate four MLLMs — three leading general-purpose models (Gemini-3, Claude Opus 4.5, GPT-5) and one medical-specialist model (MedGemma) — and find performance ranges from 78\% to 37\%, with the medical-specialist model performing worst. We also present a systematic error taxonomy with four distinct failure modes, spanning from basic geometric mistakes to high-level reasoning errors and requiring different mitigation strategies. ReXInTheWild provides the first broad-coverage benchmark for medical photograph understanding, complementing existing medical VQA datasets that focus on specialized imaging modalities and tasks (Figure 1).
\begin{figure}[!t]
    \centering
    \begin{subfigure}[t]{0.49\columnwidth}
        \centering
        \includegraphics[height=3.3cm,keepaspectratio]{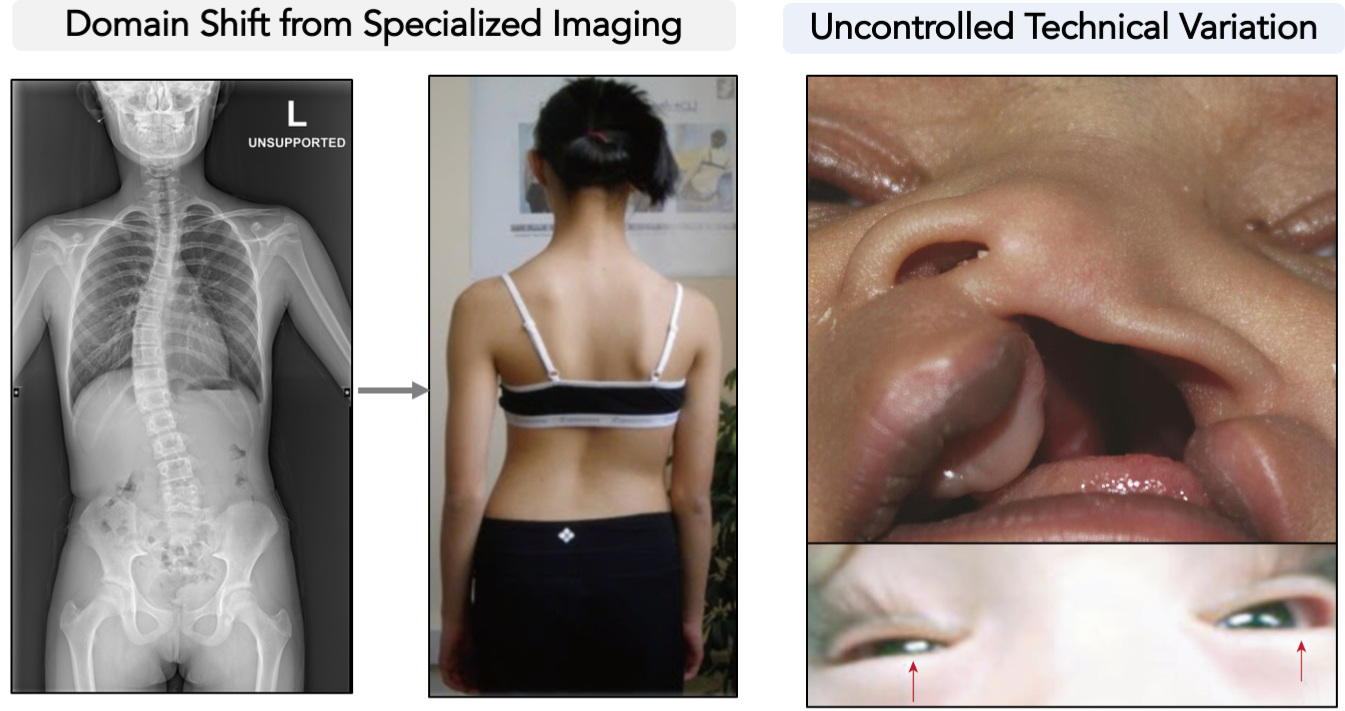}
        \caption{Challenges of medical photographs (unusual camera angles, overexposure, etc.)}
    \end{subfigure}
    \hfill
    \begin{subfigure}[t]{0.49\columnwidth}
        \centering
        \includegraphics[height=3.3cm,keepaspectratio]{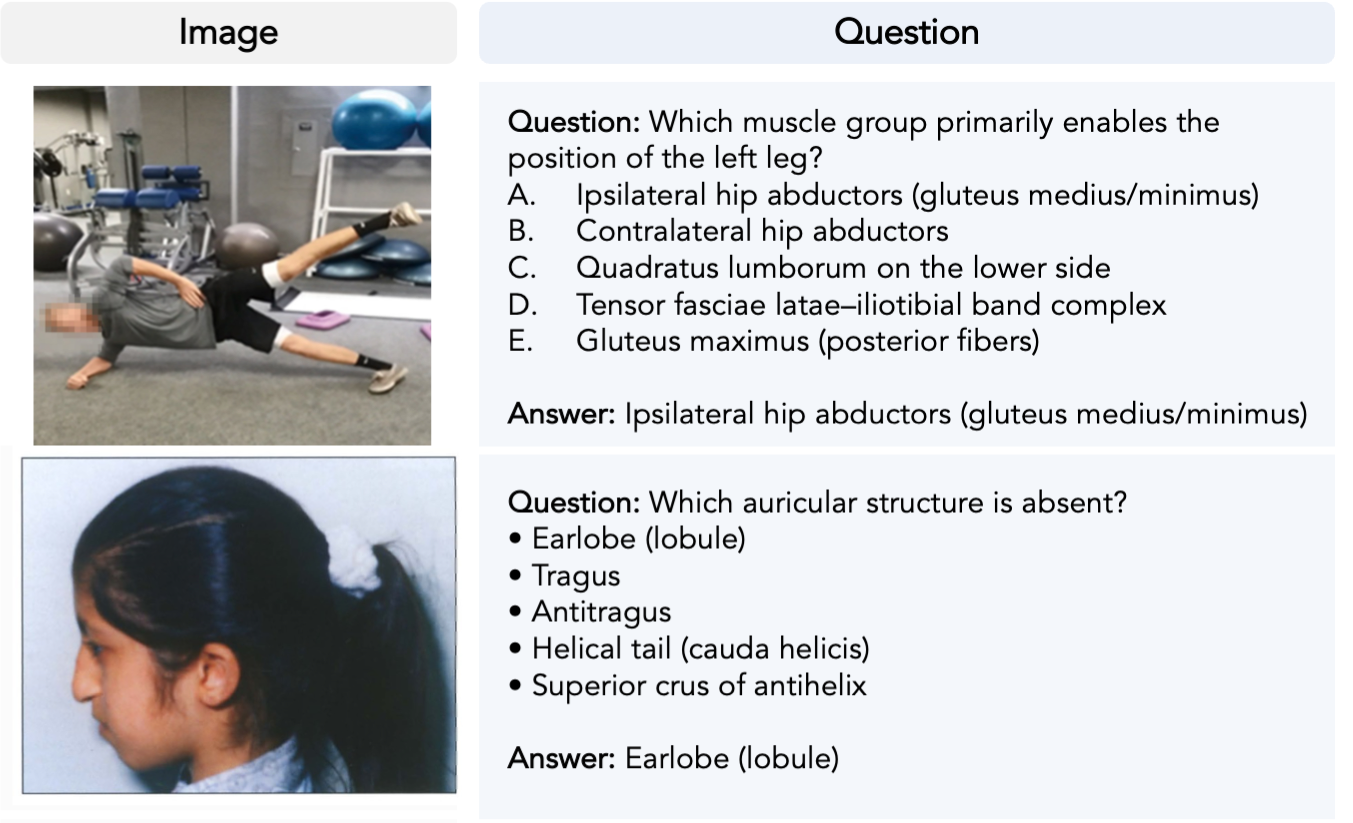}
        \caption{Example question-answer pairs from ReXInTheWild}
    \end{subfigure}

    \caption{Medical photographs pose several unique challenges at the intersection of medical reasoning and natural image interpretation. \textbf{Left}: For example, medical models face a substantial domain shift from specialized imaging modalities \cite{ho2026scoliosisradiopaedia} to photographs. In addition, photographs taken outside the clinic are likelier to have technical flaws that obscure fine-grained medical details (e.g. the abnormal lower-lid eyelashes at the bottom of ``Technical Variation"). \textbf{Right:} ReXInTheWild tests whether models can assess the medical content in diverse natural images, including both healthy subjects and patients with medical conditions.
    }
    \label{fig:teaser}
\end{figure}

\textbf{Related Benchmarks.} Numerous medical visual question-answering (VQA) benchmarks have been proposed, including VQA-RAD \cite{lau2018radiologyvqa}, PathVQA \cite{he2020pathvqa}, and SLAKE \cite{liu2021slake}, but these focus on specialized imaging modalities such as radiology and pathology. Broader multimodal evaluations such as MMMU \cite{yue2024mmmumassivemultidisciplinemultimodal} and PMC-VQA \cite{zhang2024pmcvqavisualinstructiontuning} mix a relatively small number of photographs indiscriminately with other clinical images, making it infeasible to assess performance on this particular category. Within medical photograph benchmarks, dermatology has received the most attention through datasets like DermaVQA \cite{Yim_DermaVQA_MICCAI2024} and MM-Skin \cite{zeng2025mmskinenhancingdermatologyvisionlanguage}. Smaller-scale studies have also produced fragmented datasets for other narrowly defined tasks, such as gait analysis or dental caries detection \cite{ahmed2025dentalcaries,zhou2024gaitpatternsbiomarkersvideobased,zafra2025healthgait,bandini2021facialmotion}. ReXInTheWild is, to our knowledge, the first benchmark to evaluate vision–language models across multiple clinical specialties using ordinary photographs.

\section{Methods}

\subsection{Image Collection}

We sourced images from the Biomedica dataset \cite{lozano2025biomedicaopenbiomedicalimagecaption}, which noisily clusters PubMed Central images by topic. We randomly sampled 32,982 images that were licensed for noncommercial use and labeled as ``Natural Images''.  We then performed a two-step filtering process using GPT-5 to remove unsuitable images. First, we filtered these images based on their captions by excluding captions that described subject matter other than people or body parts, such as images of landscapes or X-rays. Second, we performed a filtering pass using the images themselves and retained images that visibly depicted people or body parts in ordinary, non-clinical settings. We found that most images in the ``Natural Images" cluster were a poor fit for this dataset and fewer than 5\% of sampled images met our criteria (Figure 2a).

\subsection{Question Generation}

For each valid photograph, we used GPT-5 to generate five candidate questions, each with 3--5 multiple-choice answers. We prompted GPT-5 to produce questions that were visually and medically challenging. Additionally, we instructed the model to skip images that lacked medical content or primarily focused on dermatological topics, as those are best covered by existing datasets. 543 images remained after this step (Figure 2b).
\begin{figure}
    \centering
    \includegraphics[width=\columnwidth]{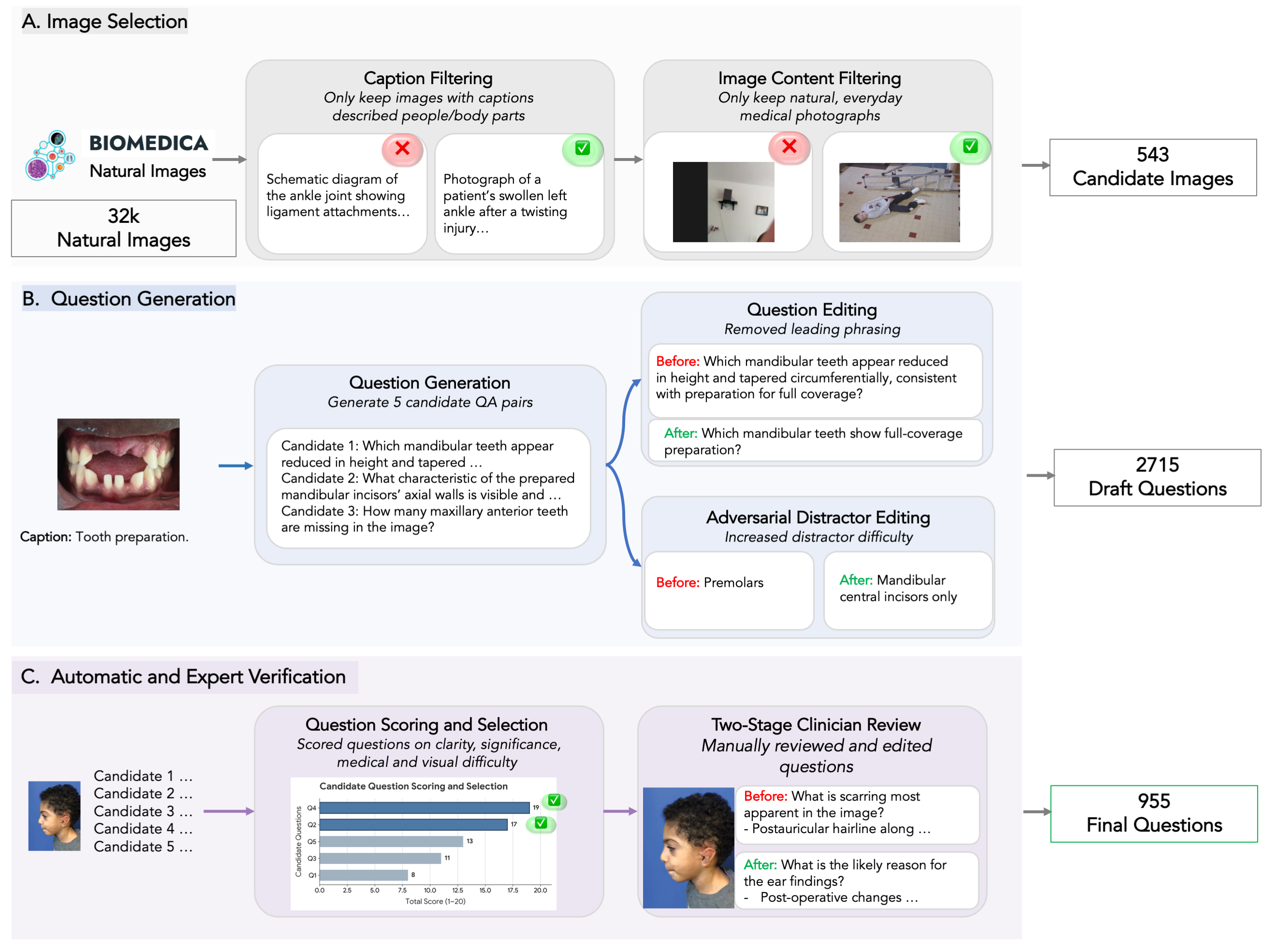}
    \caption{The ReXInTheWild benchmark construction pipeline. Stage A: ``Image Selection" filters PubMed Central images by caption and visual content to identify suitable medical photographs. Stage B: ``Question Generation" uses GPT-5 to produce candidate questions and applies automated editing to remove leading cues and refine distractors. Stage C: ``Automatic and Expert Verification" automatically scores clarity and difficulty and routes high-scoring questions for a two-stage expert review process.}
    \label{fig:fig2}
\end{figure}
Next, we performed two automatic editing passes. The first pass increased question difficulty by removing image descriptions or other leading phrases that could disclose the answer (e.g., ``Based on the location and length of the linear incision on the mid-thigh, which procedure is this scar most consistent with?'' $\rightarrow$ ``Which procedure is this scar most consistent with?''). The second pass increased question difficulty by refining the answer choices. The prompt suggested strategies for choosing convincing distractors, such as offering ``the image is normal'' when images contained subtle abnormalities.

After generating five candidate questions, we selected and manually reviewed two questions per image. To identify the most promising candidates, we used GPT-5 to score questions from 1--5 on four dimensions (clarity, medical relevance, medical difficulty, and visual difficulty) and then summed scores across dimensions.

We then selected the two top-scoring questions for each image, identifying 1086 questions for manual review. We conducted a two-stage manual review process by clinicians with different fields of expertise. In both stages, clinicians prioritized clarity and clinical relevance. They also ensured that questions varied in difficulty and cross-referenced the original PubMed articles and medical literature to validate challenging questions. In Stage 1, each question was checked by a resident or physician, who assigned the question to one of three sets: 
\begin{itemize}
    \renewcommand{\labelitemi}{$\bullet$}
    \item $\textbf{Q\_acc}$: Question-answer pairs in this set were acceptable as-is.
    \item $\textbf{Q\_rew}$: Question-answer pairs in this set needed to be rewritten to ensure clarity and clinical relevance. Clinicians could propose entirely new question topics, as long as they remained distinct from the other question on an image.
    \item $\textbf{Q\_del}$: Question-answer pairs in this set needed to be deleted outright, generally because the image was unsuitable or unable to support two distinct, clinically appropriate questions.
\end{itemize}

In Stage 2, each question and its first-stage categorization were reviewed by an experienced clinician with 6+ years of experience. They made a final decision on how to categorize each question. 645 questions were placed in $Q_{acc}$, 310 were placed in $Q_{rew}$, and 131 were placed in $Q_{del}$. This process resulted in 955 final questions based on 484 images (Figure 2c). To enable fine-grained analysis, each remaining image--question pair was tagged with one of the following seven categories: Trunk \& Extremities (39\%), Head \& Neck (26\%), Eyes (15\%), Mouth \& Jaws (12\%), Skin \& Hair (4\%), Surgical \& Procedural (3\%), and Other (1\%).

\subsection{Answer Generation}

GPT-5 responses were generated with \textbf{temperature} = 1 and \textbf{top\_p} = 1, 
using the ``2025-08-07'' version of the model and the ``2024-12-01-preview" version of the API. 
Claude Opus 4.5 responses were generated with \textbf{temperature} = 1 and 
\textbf{max\_tokens} = 512, using the ``claude-opus-4-5-20251101'' version. 
Gemini-3 responses were generated with \textbf{temperature} = 1, 
\textbf{top\_p} = 0.95, \textbf{top\_k} = 64, and \textbf{thinking} = True, 
using the ``Gemini-3-Pro-Preview'' version. 
MedGemma responses were generated in non-sampling mode with 
\textbf{num\_beams} = 1, using ``MedGemma-4B-IT'' from HuggingFace’s 
Transformers library \cite{wolf2020huggingfacestransformersstateoftheartnatural}.
We instructed all models to interpret ``left" and ``right" from the patient's perspective unless specifically instructed otherwise. Answer choices for each question were shuffled. We computed 95\% confidence intervals with the Wilson score method, using the statsmodels package \cite{seabold2010statsmodels}.

\section{Results}

\subsection{Model Performance}

\begin{figure}
    \centering
    \includegraphics[width=\columnwidth]{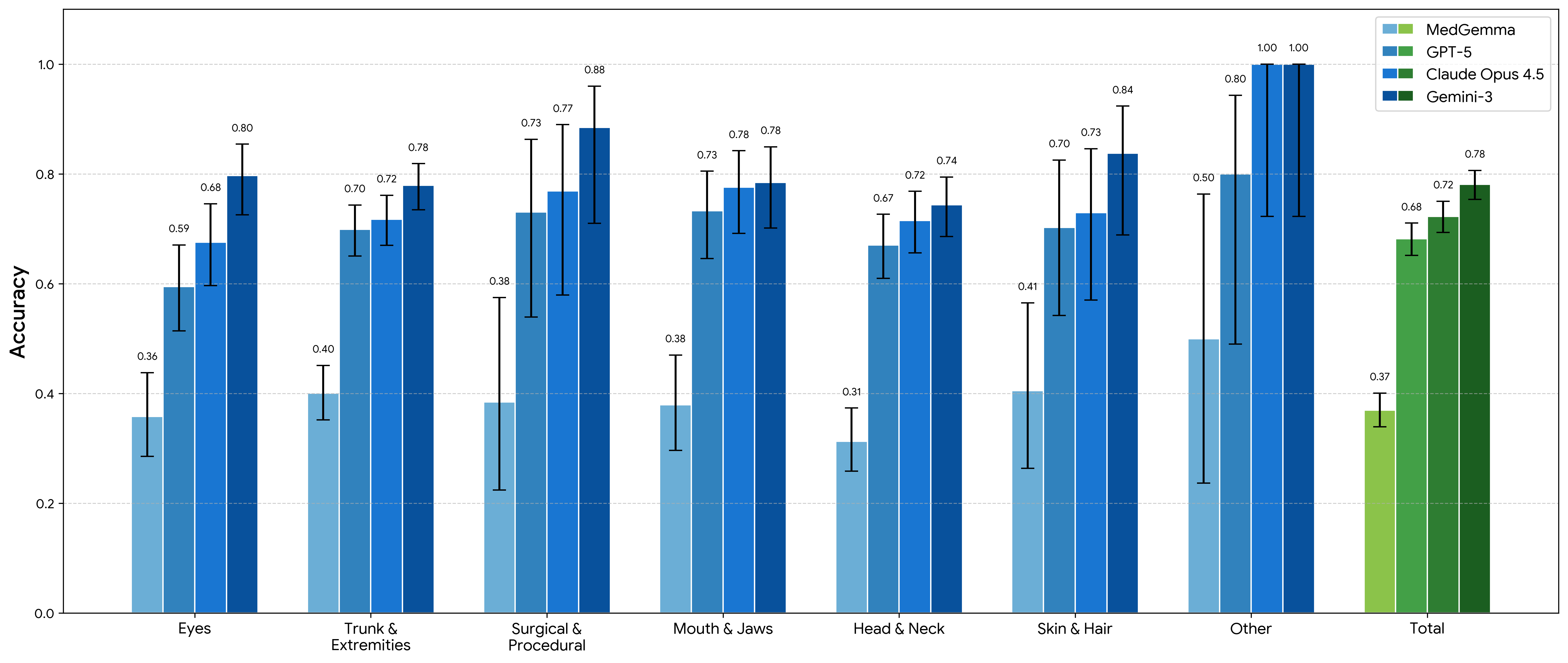}
    \caption{Large general-purpose models, especially Gemini-3, outperformed MedGemma, a smaller medical MLLM. Performance varied across medical categories, though model rankings generally remained consistent within each category.}
    \label{fig:fig5}
\end{figure}
We observed considerable variation in performance across models and topics (Figure 3). Gemini-3 achieved the strongest performance with 78\% accuracy (95\% CI: 0.75, 0.81) across all questions, followed by Claude Opus 4.5 with 72\% accuracy (95\% CI: 0.69, 0.75) and GPT-5 with 68\% accuracy (95\% CI: 0.65, 0.71). Surprisingly, MedGemma only achieved 37\% accuracy (95\% CI: 0.34, 0.40), despite being explicitly trained for medical vision-language tasks. Model rankings remained consistent across categories, and we noticed commonalities in their error patterns. GPT-5 and Claude Opus 4.5 both performed worst on ``Eyes" questions, with accuracies of 59\% (95\% CI: 0.51, 0.67) and 67\% (95\% CI: 0.60, 0.75) respectively. ``Head \& Neck" was MedGemma and Gemini-3's worst category, resulting in accuracies of 31\% (95\% CI: 0.26, 0.37) and 74\% (95\% CI: 0.69, 0.79) respectively. It was also GPT-5 and Claude Opus 4.5's second-worst category, with accuracies of 67\% (95\% CI: 0.61, 0.73) and 71\% (95\% CI: 0.66, 0.77).

\subsection{Error Analysis}
Current models, particularly Gemini-3, Claude Opus 4.5 and GPT-5, frequently showed sophisticated medical reasoning. However, we also noted gaps in both natural image interpretation and medical knowledge. We describe and quantify four error types seen across all models, in approximate order of their levels of abstraction (Figure 4):
\begin{itemize}
    \renewcommand{\labelitemi}{$\bullet$}

    \item \textbf{Measurement/Geometry Errors:} Models made relatively basic geometric errors. For example, they regularly failed to differentiate between the left and right sides of the body and between flexed and extended joints.
    \item \textbf{Localization Errors:} Models made a variety of localization errors, ranging from fine-grained medical errors (e.g. confusing the helical rim and antihelical fold) to large-scale natural image errors (e.g. not noticing that a person's feet were on the ground).
    \item \textbf{Characterization Errors:} Models failed to qualitatively describe anatomical features, incorrectly assessing their color, shape or texture. These errors can occur because of poor natural image interpretation or a lack of medical knowledge, as medical language relies heavily on specialized descriptors.
    \item \textbf{Causality Errors:} Models made logical errors, failing to understand cause-and-effect or guess the reasoning behind actions. High-level errors can easily be triggered by other, lower-level errors. For example, models frequently struggled to describe how a person's muscles would be involved in or affected by a particular pose; these mistakes may arise from theoretical difficulties with biomechanics as well as a concrete failure to recognize limb configurations.
\end{itemize}

\begin{figure}
    \centering
    \includegraphics[width=\columnwidth]{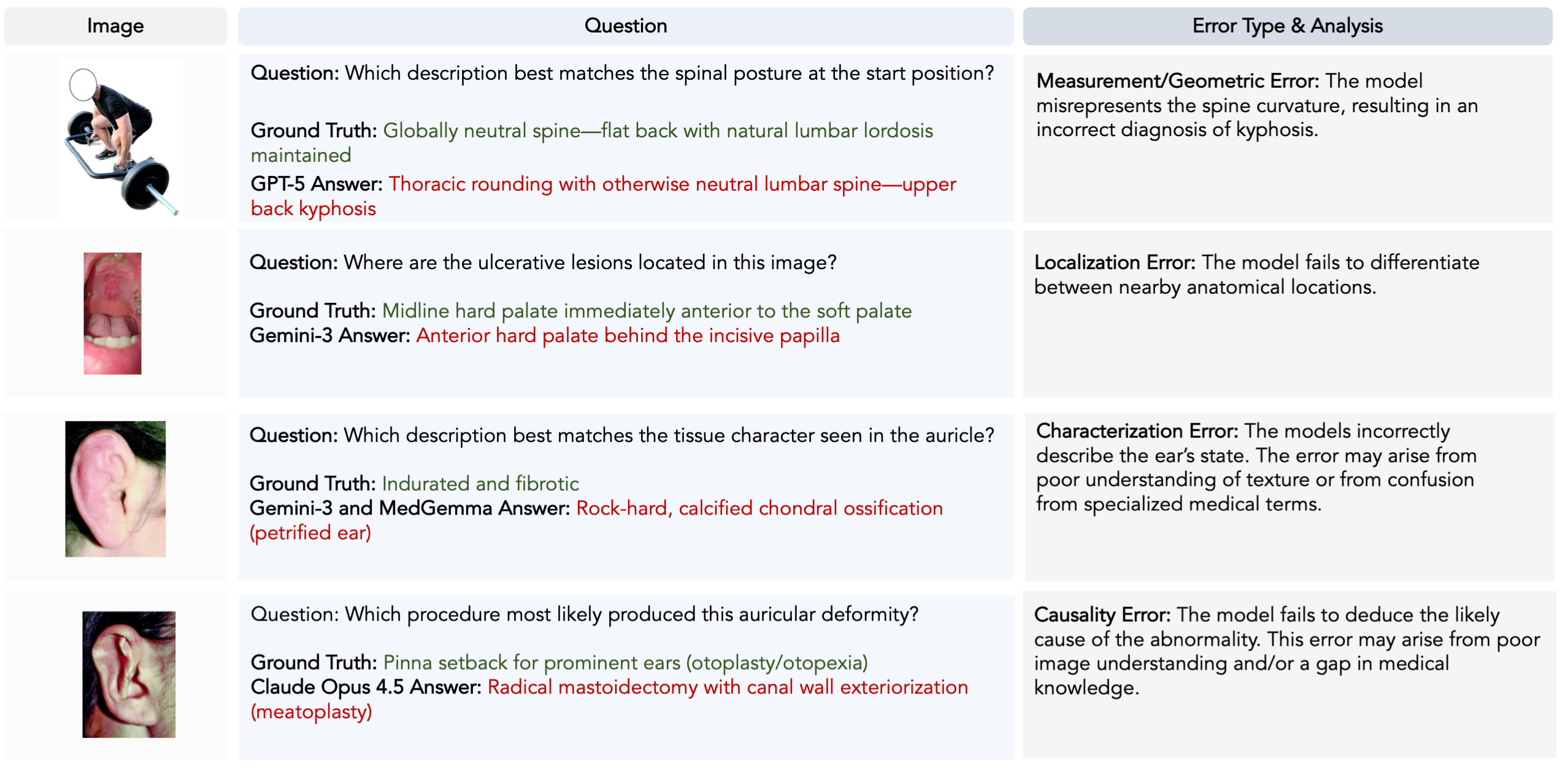}
    \caption{Failure modes range from low-level errors in basic image interpretation to high-level failures in causal reasoning. Notably, models made errors when describing normal physiology (top), not only when assessing medical abnormalities.}
    \label{fig:error_examples.png}
\end{figure}
To estimate the frequency of these errors, we sampled 50 errors made by each of the three top-performing models: Gemini-3, Claude Opus 4.5, and GPT-5. An experienced clinician classified these 150 errors into our four proposed types, marking multiple error types if they were all likely to be present (Table 1). We found that 99\% of sampled errors fell into at least one category while only 17\% belonged to multiple categories, indicating that our four categories provided broad coverage of several distinct failure modes. Errors of all levels were widespread, with geometric, localization, and characterization errors each affecting 30\%+ of cases. Causality errors were the least common type (15\%); we noted that the majority of causality errors co-occurred with another error type, suggesting that high-level errors tend to be ``built on" lower-level mistakes.

\begin{table}[ht]
\caption{Error type fractions by model and averaged across models. Our categories provide broad coverage, with 99\% of errors belonging to at least one category. They also capture relatively distinct failure modes, with only 17\% of errors falling into multiple categories.  Errors of all levels occurred frequently, indicating that current models still make basic geometric mistakes as well as sophisticated medical errors.}

\centering
\fontsize{8pt}{10pt}\selectfont
\setlength{\tabcolsep}{4.5pt} 

\begin{tabular}{lcccccc}
\toprule
Model & Geometric & Localization & Characterization & Causality & 1+ Types & 2+ Types \\
\midrule
Claude & 0.28 & 0.24 & 0.38 & 0.16 & 0.98 & 0.08 \\
Gemini & 0.34 & 0.36 & 0.36 & 0.18 & 1.00 & 0.22 \\
GPT    & 0.28 & 0.40 & 0.40 & 0.10 & 1.00 & 0.18 \\
\midrule
\textbf{Mean} & \textbf{0.30} & \textbf{0.33} & \textbf{0.38} & \textbf{0.15} & \textbf{0.99} & \textbf{0.16} \\
\bottomrule
\end{tabular}

\normalsize
\setlength{\tabcolsep}{6pt} 
\label{tab:error_fractions}
\end{table}
\section{Discussion}

ReXInTheWild represents the first comprehensive benchmark for medical photographs: natural images that may arise in non-clinical settings (``in the wild") yet contain medically relevant content. Unlike traditional medical modalities requiring specialized scanners, these photographs are easy for patients to produce and share and can capture relevant information for many specialties, making them widely useful both in and out of clinical settings. Given the rapid progress of MLLMs in both medicine and natural image interpretation, patients and clinicians alike may expect models to interpret these images accurately.

However, our benchmark reveals substantial room for improvement. The strongest model we tested, Gemini-3, reached only 78\% accuracy, while MedGemma, the model with the most targeted medical training, performed surprisingly poorly with 37\% accuracy. This discrepancy may arise from differences in model size and inference-time compute, but it also suggests that comprehension of medical photographs may benefit from heavy exposure to natural image distributions, not just curated medical images and text corpora. 

Our error analysis reveals a diverse range of failure modes. We observed that models struggled to comprehend not only medical abnormalities but also the physiological underpinnings of everyday movements and poses. As a result, models are at risk of mishandling common inquiries about exercise, posture, and physical therapy, as well as rarer specialized questions. Expert review showed that failures could be classified into four distinct categories, which likely require different technical mitigation strategies. Low-level geometric errors (e.g. incorrect measurement of angles) and localization errors could potentially be fixed simply by equipping MLLMs with fine-grained keypoint or object detection tools. Mid-level errors, such as incorrect qualitative descriptions, may require search and lookup tools that define medical terms and highlight similar cases. High-level errors in causal reasoning may be addressed through additional inference-time computation (e.g. chain of thought), as well as through resolving underlying low-level errors. Overall, our findings suggest that future MLLMs should not be fine-tuned on either clinical data or natural images in isolation. Instead they must blend general visual understanding with targeted medical knowledge, potentially through retrieval-augmented generation or modular tool-use frameworks.

\textbf{Limitations.} Because images were sourced from scientific articles, some were annotated, arranged in panels, or partially redacted, differing from typical photographs. The dataset also likely overrepresents long-tail conditions and severe presentations of conditions, reflecting research interests. Although we included some images without medical abnormalities, patient-generated photographs may be more likely to show common findings and mild presentations. Using GPT-5 during initial question generation may introduce bias, though this concern is mitigated by the fact that Gemini-3 and Claude Opus 4.5 both outperform it.  

Finally, some medical assessments simply should not be performed using static images: certain questions require palpation, motion, advanced imaging, or other information that a photograph alone cannot provide. These limitations underscore the complexities of medical photographs and highlight the need for future research on what MLLMs can and cannot infer from them. Future work should explore impact on patients, particularly marginalized subgroups, and ensure that models do not mislead patients about their health.

\section{Acknowledgements}
O.B. was  supported by the Biswas Family Foundation’s Transformative Computational Biology Grant in collaboration with the Milken Institute. S.E.K. was supported by a grant of the Boston-Korea Innovative Research Project through the Korea Health Industry Development Institute(KHIDI), funded by the Ministry of Health \& Welfare, Republic of Korea(Grant \# RS-2024-00403047).

\newpage

\end{document}